\pdfoutput=1

\documentclass[11pt]{article}

\usepackage[final]{acl}

\usepackage{times}
\usepackage{latexsym}

\usepackage[T1]{fontenc}

\usepackage[utf8]{inputenc}

\usepackage{microtype}

\usepackage{inconsolata}

\usepackage{amsfonts,amssymb}
\usepackage{amsmath}
\usepackage{graphicx}
\usepackage{multirow}
\usepackage{booktabs}
\usepackage{subfigure}
\usepackage{parskip}
\usepackage[export]{adjustbox}
\usepackage{subcaption}
\usepackage{enumitem}
\usepackage{algorithm}
\usepackage{algpseudocode}
\usepackage{tcolorbox}
\tcbuselibrary{breakable}
\usepackage{caption}
\usepackage{xcolor}
\usepackage{multicol}
\usepackage{comment}

\title{DKEC: Domain Knowledge Enhanced Multi-Label Classification for Diagnosis Prediction}

\author{Xueren Ge \and Satpathy Abhishek \and Ronald Dean Williams, \\
 {\bf John A. Stankovic} \and {\bf Homa Alemzadeh} \\
  University of Virginia, Charlottesville, VA 22903 USA \\
  \texttt{\{zar8jw, cqa3ym, rdw, jas9f, ha4d\} @virginia.edu}}

\begin{document}
\maketitle
\begin{abstract}

Multi-label text classification (MLTC) tasks in the medical domain often face the long-tail label distribution problem. Prior works have explored hierarchical label structures to find relevant information for few-shot classes, but mostly neglected to incorporate external knowledge from medical guidelines. This paper presents DKEC, \textbf{D}omain \textbf{K}nowledge \textbf{E}nhanced \textbf{C}lassification for diagnosis prediction with two innovations: (1) automated construction of heterogeneous knowledge graphs from external sources to capture semantic relations among diverse medical entities, (2) incorporating the heterogeneous knowledge graphs in few-shot classification using a label-wise attention mechanism. We construct DKEC using three online medical knowledge sources and evaluate it on a real-world Emergency Medical Services (EMS) dataset and a public electronic health record (EHR) dataset. Results show that DKEC outperforms the state-of-the-art label-wise attention networks and transformer models of different sizes, particularly for the few-shot classes. More importantly, it helps the smaller language models achieve comparable performance to large language models.
\end{abstract}

\section{Introduction}
Automated diagnosis prediction~\cite{ma2017dipole} is the challenging task of classifying different diseases based on a patient's EHR for applications such as treatment recommendation (e.g., selecting EMS protocols~\cite{shu2019behavior,jin2023emsassist,weerasinghe2024real}) or medical billing (e.g., assigning ICD-9 codes)~\cite{o2005measuring}. 

Diagnosis prediction based on the free-text medical notes is known as multi-label text classification (MLTC)~\cite{liu2017deep}, which is the task of assigning the most relevant labels to a text instance. MLTC is more complex than the traditional multi-class text classification because the number of possible \textit{label combinations grows exponentially} with the number of classes~\cite{chen2017ensemble}. 
Another challenge in diagnosis prediction is the imbalanced distribution of diagnoses as some medical conditions happen more frequently than others, causing a \textit{long-tail data distribution}. For example, the total number of chest pain-related reports in a real-world EMS dataset is ten times more than overdose/poisoning reports \cite{kim2021information}. 
Training on such imbalanced datasets, also called "power-law datasets" \cite{rubin2012statistical}, introduces bias in model predictions towards \textit{head} label classes while ignoring the \textit{few-shot} or \textit{tail} classes.

\begin{table*}[t!]
\centering
\setlength{\tabcolsep}{4pt} 
\resizebox{\textwidth}{!}{
\begin{tabular}{c|c|c|c|c|c}
\hline
Models & Encoder & Attention Mechanism & Knowledge Integration & Knowledge Source & Datasets \\
\hline
\cite{van2021clinical} & BERT & Self-Attention & Pre-training & Wikipedia, PubMed & MIMIC-III\\
\cite{yang2022gatortron} & MegatronBERT & Self-Attention & Pre-training & Wikipedia, PubMed & MIMIC-III\\
\cite{bolton2024biomedlm} & GPT2 & Self-Attention & Pre-training & PubMed & MedMCQA\\

\hline
\cite{mullenbach-etal-2018-explainable} & CNN & Label-wise Attention & & & MIMIC-III\\
\cite{rios2018few} & CNN & Label-wise Attention & ICD-9 hierarchy graph & ICD-9 description & MIMIC-III\\

\cite{li2020icd} & Multi-filter residual CNN & Label-wise Attention & & & MIMIC-III\\

\cite{zhou2021automatic} & Multi-filter CNN & Shared Interactive Attention & & & MIMIC-III\\
\hline

DKEC (Ours) & Multi-filter CNN, Transformers & Label-wise Attention & Heterogeneous graph & Wikipedia, MayoClinic, ODEMSA & MIMIC-III \& EMS\\

\hline
\end{tabular}%
}
\vspace{-0.1in}
\caption{Summary of previous works on diagnosis prediction.}
\label{tab:sota_summary}
\vspace{-1.25em}
\end{table*}

Most existing diagnosis prediction solutions~\cite{rasmy2021med, lee2020biobert} are task-agnostic and rely on integrating biomedical domain knowledge with transformer models in the pre-training stage. For example, COReBERT~\cite{van2021clinical} uses clinical outcome pre-training to learn relations among symptoms, risk factors and clinical outcomes by incorporating Wikipedia and PubMed knowledge bases. Recent pre-trained large language models (LLMs)~\cite{yang2022large, luo2022biogpt, bolton2024biomedlm} have demonstrated superior performance by leveraging large external clinical corpora and huge number of parameters. However, these models only incorporate uncurated knowledge in pre-training, neglect task-specific knowledge and label relations, and are costly to fine-tune and deploy on resource-constrained devices~\cite{jin2023emsassist, weerasinghe2024real}. 


To solve the class-imbalance problem in MLTC, the convolutional attention network and its variants~\cite{kim-2014-convolutional, li2020icd, liu2021effective} were proposed to extract meaningful document representations that cover different ranges of clinical text. Other works~\cite{rios2018few, wang-etal-2022-kenmesh} integrated hierarchical information by graph convolutional neural networks to select label-relevant features. Some follow-up studies~\cite{lu-etal-2020-multi, cao2020hypercore, zhou2021automatic} have also proposed to incorporate label co-occurrence graphs, along with hierarchical structures to capture label concurrent and mutual exclusive relations for ICD-9 code classification. However, most of these works neglected the potential benefits of incorporating \textit{expert knowledge from medical guidelines}. External domain knowledge can provide additional information for training with few-shot labels to compensate for data scarcity and model size or be applied as constraints in training based on \textit{label relations}.

\begin{figure*}[t!]
    \centering
    \subfigure[DKEC Pipeline]{
        \label{fig:dkec pip}
        \includegraphics[width=0.7\linewidth,height=5.7cm]{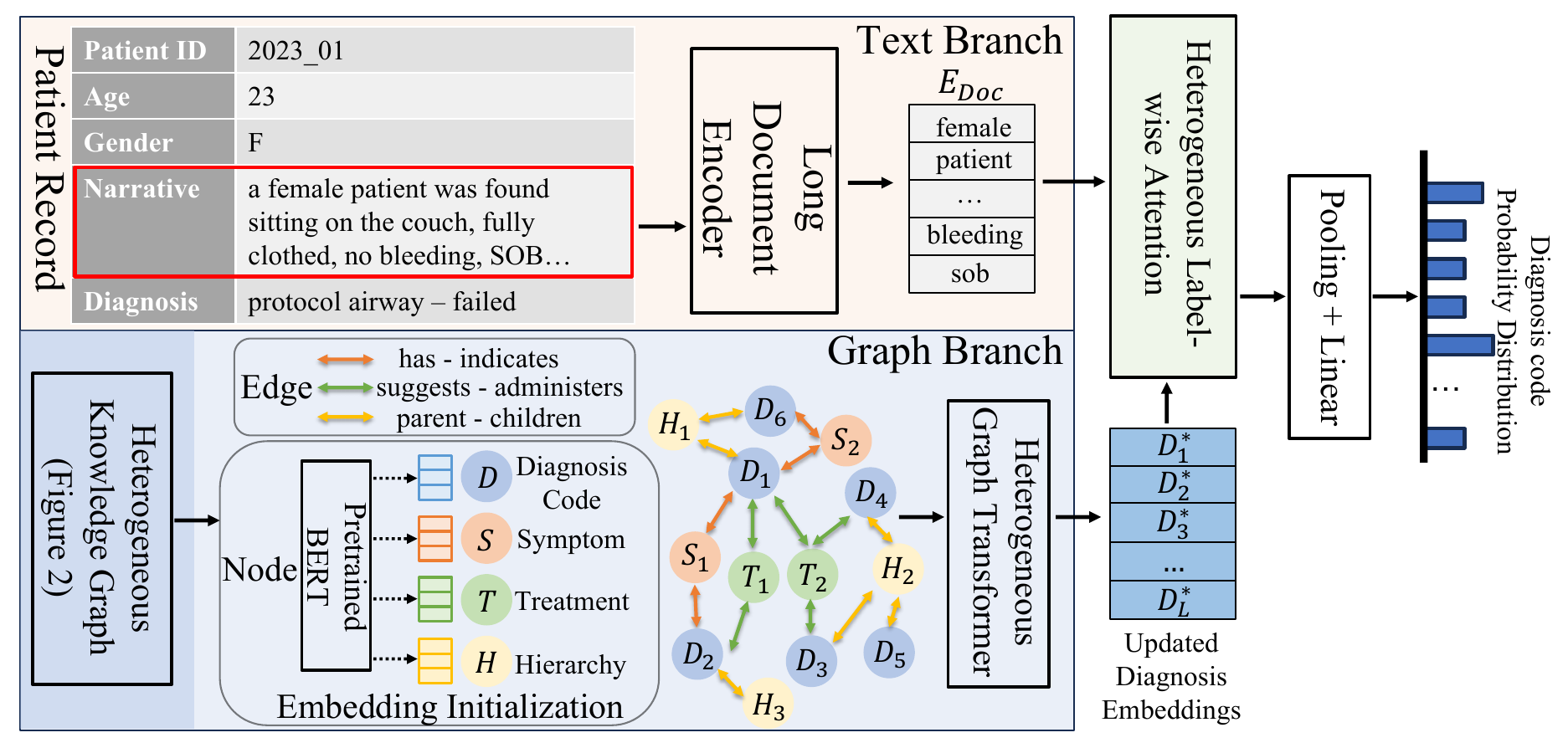}
    }\noindent
    \subfigure[HLA]{
        \label{fig:hla}
        \includegraphics[width=0.22\linewidth,height=5.5cm]{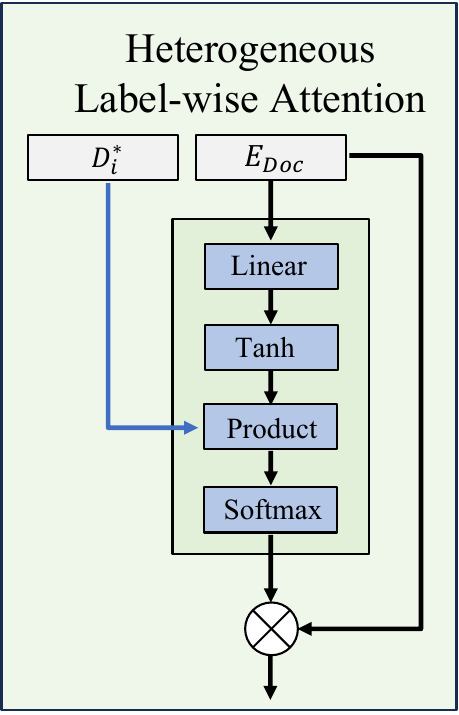}
    }
        
\vspace{-1em}
\caption{(a) DKEC Pipeline includes three main modules: a text branch to derive text embeddings, a graph branch to derive updated diagnosis embeddings, and (b) an HLA module to derive label-attentive document embeddings.}
\label{Pipeline}
\vspace{-1em}
\end{figure*}

This paper presents DKEC (Figure \ref{Pipeline}), a \textit{knowledge and data-driven} approach to class-imbalanced MLTC by (i) automated extraction of \textit{label-specific semantic relations} from \textit{online sources} and (ii) integrating them as \textit{heterogeneous knowledge graphs} with different encoders using a \textit{label-wise attention mechanism}. Our contributions are as follows:



\begin{itemize}
    \item We develop a method for automated construction of heterogeneous knowledge graphs from online sources (e.g., Wikipedia, MayoClinic, ODEMSA) that accurately captures semantic relations among diverse medical entities (e.g., symptoms and diseases, diseases and treatments), by medical entity extraction using chain-of-thought prompting with GPT-4 and UMLS medical concept normalization.   
    \item We design a heterogeneous label-wise attention mechanism based on graph transformers that captures the diagnosis co-occurrence relations based on relevant medical entities in the knowledge graph and is combined with different encoders (e.g., Multi-filter CNN, BERT) to improve multi-label classification. 
    \item We conduct extensive experiments to evaluate DKEC by applying it to language models of varying sizes using a real-world EMS dataset~\cite{kim2021information} and the MIMIC-III dataset~\cite{johnson2016mimic}. Results show that DKEC outperforms state-of-the-art by 3.7\% and 2.1\% in overall top-K recall for the EMS and MIMIC-III datasets, respectively, and enhances small language model performance in few-shot classes by 10.5\% and 6\%.
    
\end{itemize}

\section{Related Work}

\textbf{Pre-trained Transformers for Diagnosis Prediction}
One avenue explored in prior work is focused on large-scale pre-training from clinical admissions, discharge summaries, and other biomedical texts, such as BioBERT~\cite{lee2020biobert}, COReBERT~\cite{van-aken-etal-2021-clinical} and GatorTron~\cite{yang2022large} (Table~\ref{tab:sota_summary}).  Recently, it has been shown that pre-trained LLMs, including BioGPT~\cite{luo2022biogpt} and BioMedLM~\cite{bolton2024biomedlm}, can outperform general-purpose models and compete with expert-designed, domain-specific model architectures. Unlike these works, which focus on integrating external knowledge corpora for task-agnostic pre-training, we aim to incorporate task-specific knowledge and disease-related relations during the fine-tuning stage.

\textbf{Label-wise Attention Networks}
Another line of research has focused on developing attention mechanisms to select the most relevant clinical segments for each label (see Table~\ref{tab:sota_summary}). CAML~\cite{mullenbach-etal-2018-explainable} was the first that proposed to integrate the semantic meanings of the labels by assigning label-wise attention weights to medical text. In~\cite{rios2018few, ijcai2020-461-vu}, the hierarchical structure of labels was modeled and further concatenated into text features for classification. Recent studies have proposed different modules, including multiple graph aggregation~\cite{lu-etal-2020-multi}, interactive shared representation network~\cite{zhou2021automatic}, and hyperbolic and co-graph representation learning module~\cite{cao2020hypercore} to capture label co-occurrence along with label hierarchy for ICD code classification. However, these works ignore the domain knowledge from other sources (e.g., medical guidelines), which can provide additional information for training with rare classes and compensate for data scarcity. Also, most of them only focused on ICD coding using convolutional neural networks (CNNs) in the MIMIC-III dataset and showed low performance for diagnosis prediction~\cite{mullenbach-etal-2018-explainable}. 

\textbf{Biomedical Knowledge Graph Construction}
Knowledge graphs provide an efficient way to organize and access the expanding biomedical knowledge. Most existing works~\cite{harnoune2021bert, xu2020building} utilize BERT models to construct biomedical knowledge graphs through named entity recognition and relation extraction. However, BERT is limited by its capacity to process only a fixed number of tokens and is trained for pre-determined named-entity classes, making it unsuitable for long biomedical literature with an open set of named-entities. Recent research~\cite{agrawal-etal-2022-large, arsenyan2023large, goel2023llms, hu2024improving} shows that LLMs possess excellent zero-shot information extraction capabilities, thus can be suitable for constructing knowledge graphs.


\section{Heterogeneous Knowledge Graph Construction}
Our goal is to construct a heterogeneous knowledge graph $G$, which for every disease diagnosis code $D_k$ in a set of \textit{Diagnosis Codes} $D\colon\{D_k\}_{k=1}^{L}$ ($L$ is the total number of diseases), represents the corresponding sets of medical concepts such as \textit{Signs and Symptoms} $S\colon\{S_k\}_{k=1}^{|S|}$, \textit{Treatments} $T\colon\{T_k\}_{k=1}^{|T|}$ and \textit{Hierarchy} $\colon\{H_k\}_{k=1}^{|H|}$ that have semantic relations with $D_k$. As shown in Figure~\ref{Pipeline}, the heterogeneous graph of medical concepts is constructed as $G = \left(N, E\right)$, with $N$ as the set of nodes and $E$ as the set of edges. There are four different types of nodes in the graph, diagnosis codes $D$, signs and symptoms $S$, treatment $T$, and hierarchy $H$, and three types of bidirectional edges,
has/indicates $\mathop{E_{DS}}\limits ^{\longleftrightarrow}$ between $D$ and $S$, suggests/administers $\mathop{E_{DT}}\limits ^{\longleftrightarrow}$ between $D$ and $T$, and children/parent $\mathop{E_{DH}}\limits ^{\longleftrightarrow}$ between $D$ and $H$. For example, the “Injury - Crush Syndrome” diagnosis code $D_i$ is connected to the signs and symptom “muscle mass” $S_j$ using an edge of type “has/indicates” $\mathop{E_{DS}}\limits^{\longleftrightarrow}$. 
"Injury – Head" ($D_{j}$) and $D_i$ are the children of ($\mathop{E_{DH}}\limits^{\longleftrightarrow}$) node "Adult Trauma Emergencies".

Next, we present a systematic method for automated extraction of disease-relevant medical entities from external knowledge sources and mapping them into normalized concepts for unique representation in the graph. For every diagnosis code in a training dataset, we extract the relevant symptoms and treatments for the disease textual descriptions in online sources and generate the triplets $<D_k, relation, T_k/S_k>$. The hierarchy information is given by the label coding in each dataset.



\begin{figure*}[ht!]
\centering
\includegraphics[width=0.9\textwidth,height=6cm]{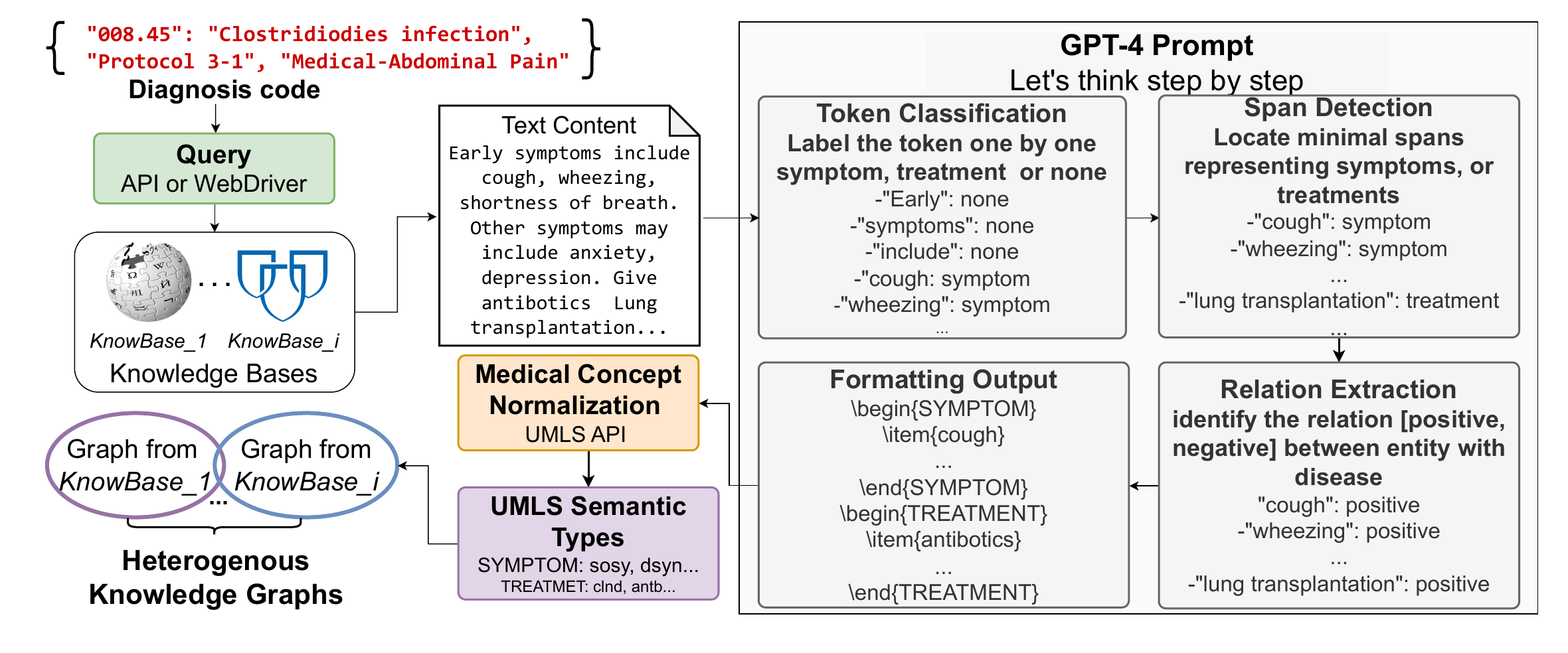}
\vspace{-1.25em}
\caption{Knowledge Graph Construction}
\label{KG}
\vspace{-1em}
\end{figure*}

\subsection{Medical Entity Extraction}
Online knowledge bases (KBs) are heterogeneous and contain different information on diseases, so we utilize multiple KBs for medical entity extraction (see Figure~\ref{KG}). 
For a disease diagnosis code $D_k$, we use the textual description of the disease as the search term to query the KBs. This is done using API calls for KBs with readily available API endpoints and a WebDriver for those without APIs. We then extract the content from the first page of the search results and identify the symptoms and treatments from the text. We evaluated different medical entity extraction methods and used prompting \textit{gpt-4-1106-preview}~\cite{achiam2023gpt} via one-shot chain-of-thought (one-shot CoT)~\cite{NEURIPS2022_9d560961} as it showed the best performance (see Sections~\ref{kb_description} and~\ref{KG_quality_eval}). As shown in Figure~\ref{KG}, one-shot CoT prompts decompose the task into three sub-tasks and ask GPT-4 to think step by step:\\ 
\textbf{Token Classification} asks GPT-4 to label each token in the text by symptom, treatment or none.\\
\textbf{Span Detection} is the task to locate the minimal span representation of a medical phrase, which is necessary for symptoms like "shortness of breath." This can also help to refine the extracted medical entities better by removing irrelevant modifiers. These two steps are important to prevent GPT-4 from rephrasing medical entities or hallucinations.\\
\textbf{Relation Extraction} asks GPT-4 to check if the extracted medical phrases from the text are related to the disease to avoid scenarios like negation.


\subsection{Medical Entity Representation}
For the unique representation of nodes in the graph, we first do \textit{medical concept normalization}. We use the Unified Medical Language System (UMLS) REST API~\cite{bodenreider2004unified} to map the extracted entities presented in different semantic variations to normalized medical concepts. For instance, the entities "fever", "high temperature", and "burning up" can all be mapped to the same UMLS concept "Fever" with Concept Unique Identifier (CUI) C0015967. Specifically, we define semantic type sets for symptoms and treatments ~\cite{agrawal-etal-2022-large, parwez2018biomedical} and for each entity take the first returned normalized concept with the right semantic type from the API (\hyperref[sec:umls_norm]{Appendix A.1}).  

Then, we generate an \textit{initial node embedding} for each node in the graph by applying a pre-trained BERT model to the node's textual description (see Figure~\ref{fig:dkec pip}) and summing up hidden states in the last four layers for the trade-off between memory and performance~\cite{kenton2019bert}.



\section{Knowledge-Enhanced Classification} Figure~\ref{fig:dkec pip} shows the overall DKEC pipeline, consisting of (i) a text branch that extracts features from the input EHR notes using a long document encoder (ii) a graph branch that takes the heterogeneous knowledge graph $G$ as input and incorporates it for label co-occurrence extraction and multi-label classification via a Heterogeneous Label-wise Attention (HLA) mechanism.
\subsection{Long Document Encoder}
Given that our input is the long text $Doc$ in EHR notes, we need models that can handle temporal sequences and medical terminologies for feature extraction. Prior work \cite{ji2022unified} shows CNNs have superior performance on clinical document classification and pre-trained transformers are limited to encoding a maximum sequence length of 512. To make a fair comparison, we apply DKEC to different state-of-the-art encoders, including multi-filter CNNs and pre-trained transformers.

For multi-filter CNN, similar to work~\cite{zhou2021automatic}, we first map the words in the input text into the low-dimensional word embedding space, then concatenate the convolutional representation from kernel set with different sizes to generate document features $\mathbf{E}_{Doc}$. For BERT models, similar to work~\cite{ji2021does}, we chunk the long document into shorter texts and concatenate all the chunked text features from the hidden states in the last layer of the BERT to generate document features $\mathbf{E}_{Doc}$.


\subsection{Heterogeneous Label-wise Attention}

We use the heterogeneous graph transformer (HGT)~\cite{hu2020heterogeneous} as the graph model and add another linear layer on the top of HGT's output to derive final embeddings for all the labels (diagnosis nodes). The input of HGT is the initial node embeddings and the medical concept relations, and the output is the updated node embeddings, from which we only use the updated diagnosis embeddings for HLA construction. In the feed-forward phase of the HGT model, a diagnosis node $D_k$ aggregates information from neighboring medical concept nodes $S_k$, $T_k$ by giving different weights to update itself as $\mathbf{D}^{\star}_{k}$. We denote the set of updated diagnosis embeddings from HGT as  $\mathbf{D}^{\star}\colon\left\{\mathbf{D}^{\star}_{k}\right\}_{k=1}^{L}$,
\begin{equation} 
    \mathbf{D}^{\star} = \mbox{Linear}(\mbox{HGT}(G))
\end{equation}
\vspace{-0.02in}
where $\mathbf{D}^{\star} \in \mathbb{R}^{L \times \delta}$ is the label representation which incorporates knowledge from diverse medical entities and captures co-occurrence relations in diagnosis codes and $\delta$ is a hyper-parameter indicating the dimension of hidden states. 
We then design an HLA to combine knowledge from each label representation $\mathbf{D}^{\star}_{k} \in \mathbf{D}^{\star}$ with text representation $\mathbf{E}_{t}$, by having the labels assign different weights for each token in the document representation. The label-wise attention vector is constructed as: 
\vspace{-0.05in}
\begin{equation}
    \mathbf{a}_{Doc, k} = \mbox{softmax}(\mbox{tanh}(\mathbf{W}_0 \mathbf{E}_{Doc} + \mathbf{b}_0) \mathbf{D}^{\star}_{k})
    \vspace{-0.2em}
\end{equation}
\begin{equation}
    \mathbf{A}_{Doc} = [
    \mathbf{a}_{Doc,1} \cdots \mathbf{a}_{Doc, k} \cdots \mathbf{a}_{Doc, L}
    ]^T
\end{equation}
\vspace{-0.02in}
where $\mathbf{W}_0$ and $\mathbf{b}_0$ are respectively the weight and bias of a linear layer to match the size of hidden dimensions in the document representation with the size of label representation and $\mathbf{a}_{Doc, k} \in \mathbb{R}^{\text{Seq}\times n}$ measures how much weight the $k$th label assigns to each token in document $Doc$. Finally, we combine all attention vectors $\mathbf{a}_{Doc, k}$ of a document $Doc$ for all $L$ labels to have $\mathbf{A}_{Doc} \in \mathbb{R}^{L \times (\text{Seq}\times n)}$, then the label-wise text representation $\mathbf{E}_{Doc}^{attn} \in \mathbb{R}^{L \times \delta}$ is generated as follows,
\begin{equation}
    \mathbf{E}_{Doc}^{attn} = \mathbf{A}_{Doc} \mathbf{E}_{Doc}
\end{equation}
\vspace{-0.02in}
which measures how informative medical text $t$ is for different labels.

\subsection{Classification}
The classification layer aims to find the most relevant label $\hat{y}_{t}$ to the input document $Doc$. We add another pooling layer for the features obtained from HLA ($\mathbf{E}_{Doc}^{attn}$) 
before the linear layer to reduce memory usage. The final prediction based on probabilities for each class $\hat{y}_{Doc} \in \mathbb{R}^{L}$ is achieved after the linear layer:
\vspace{-0.05in}
\begin{equation}
    \hat{y}_{Doc} = \mbox{Linear}(\mbox{Pooling}(\mathbf{E}_{Doc}^{attn}))
\end{equation}

The binary cross-entropy loss is applied to measure the distance between each predicted label $\hat{y}_{Doc}$ and ground-truth $y_{Doc}$.
\vspace{-0.05in}
\begin{equation}
\begin{aligned}
    \mathcal{L} = -\sum^{|Doc|}_{Doc=1}\sum^{L}_{l=1}(y_{Doc,l} \log(\hat{y}_{Doc,l}) +\\ (1 - y_{Doc,l}) \log(1 - \hat{y}_{Doc, l}))
\end{aligned}
\vspace{-1em}
\end{equation}

\begin{table*}
\centering
\small
\begin{tabular}{|c|cc|cc|cc|}
\hline
\multicolumn{1}{|c|}{}& \multicolumn{2}{|c|}{Wikipedia (50 ICD-9 codes)} & \multicolumn{2}{|c|}{Mayo Clinic (50 ICD-9 codes)} & \multicolumn{2}{|c|}{ODEMSA (43 EMS protocols)}\\
\cline{2-7}
\multicolumn{1}{|c|}{\textit{wo/w NORM}} & Symptom & Treatment & Symptom & Treatment & Symptom & Treatment\\
\cline{1-7}
MetaMap & 47.62 / 51.53 & 34.66 / 41.95 & 44.83 / 49.12 & 41.82 / 46.44 & 41.34 / 43.61 & 39.20 / 41.95\\
cTAKES & 48.74 / 52.58 & 36.01 / 43.35 & 42.60 / 46.67 & 39.67 / 45.35 & 38.02 / 42.47 & 48.96 / 52.31\\
ScispaCy & 52.79 / 55.57 & 41.73 / 49.71 & 46.54 / 50.43 & 45.94 / 50.89 & 44.39 / 47.69 & 35.88 / 38.82\\
zero-shot GPT-4 & 51.99 / 58.77 & 17.93 / 32.13 & 52.98 / 63.37 & 26.16 / 36.48 & 76.07 / 79.72 & 10.17 / 23.50\\
one-shot CoT GPT-4 & \textbf{84.63 / 86.57} & \textbf{85.70 / 89.12} & \textbf{82.03 / 86.72} & \textbf{90.43 / 93.90} & \textbf{86.96 / 91.01} & \textbf{86.48 / 88.92} \\
\hline
\end{tabular}
\vspace{-0.1in}
\caption{Comparison with baselines on three knowledge bases. \textit{wo/w NORM} means the micro F1-scores are measured before/after medical entity normalization. The best results are highlighted in \textbf{bold}.}
\label{KG evaluation}
\vspace{-1em}
\end{table*}

\section{Experiments}
We conduct extensive experiments to evaluate DKEC by applying it to different baseline language models and comparing its performance to state-of-the-art (SOTA) diagnosis prediction methods. In our experiments, we aim to answer three research questions:\\
\textbf{RQ1:} Can DKEC improve MLTC performance for class-imbalanced datasets?\\
\textbf{RQ2:} How does DKEC perform when applied to language models with varying sizes?\\
\textbf{RQ3:} How does DKEC perform with scaling label sizes?

\begin{table}
\centering
\small
\setlength{\tabcolsep}{4pt} 
\begin{tabular}{|c|ccc|ccc|}
\hline
& ~ & ~ & ~ & \multicolumn{3}{c|}{$N_{l}$} \\
Dataset & $N_{train}$ & $N_{val}$ & $N_{test}$ & H & M & T\\
\hline
EMS & 2787  & 314 & 1316 & 10 & 21 & 12 \\
MIMIC-III & 47413 & 1627 & 3363 & 494 & 1038 & 2205 \\
\hline
\end{tabular}%
\vspace{-0.5em}
\caption{Dataset statistics, $N_{train}$: number of training instances, $N_{val}$: number of validation instances, $N_{test}$: number of test instances, $N_{l}$: number of labels in total.}
\label{tab:data statistics}
\vspace{-1.75em}
\end{table}

\begin{table*}[htbp]
\small
\centering
\setlength{\tabcolsep}{2 pt} 
\begin{tabular}{|c|c|cc|cc|cc|cccc|} 
\hline
\multicolumn{2}{|c|}{} & \multicolumn{2}{c|}{Head Labels}& \multicolumn{2}{c|}{Middle Labels} & \multicolumn{2}{c|}{Tail Labels} & \multicolumn{4}{c|}{Overall}\\
\cline{3-12}
\multicolumn{2}{|c|}{} & P@1 & R@1 & P@1 & R@1 & P@1 & R@1 & miF & maF & P@1 & R@1\\
\cline{1-12}
\multirow{8}{*}{\rotatebox{90}{EMS}} 
& CAML & $78.6_{\pm 1.3}$ & $77.7_{\pm 1.3}$ & $33.0_{\pm 0.5}$ & $32.6_{\pm 0.6}$ & $22.7_{\pm 4.5}$ & $22.7_{\pm 4.5}$ & $63.7_{\pm 1.2}$ & $22.4_{\pm 1.3}$ & $65.0_{\pm 1.6}$ & $63.5_{\pm 1.5}$ \\
~ & ZAGCNN & $83.0_{\pm 1.0}$ & $82.0_{\pm 1.0}$ & ${47.0}_{\pm 1.0}$ & ${46.2}_{\pm 0.7}$ & ${37.9}_{\pm 7.7}$ & ${37.9}_{\pm 7.7}$ & $64.8_{\pm 1.1}$ & ${28.3}_{\pm 2.0}$ & ${69.6}_{\pm 0.7}$ & ${68.1}_{\pm 0.6}$ \\
~ & MultiResCNN & ${84.3}_{\pm 0.2}$ & ${83.2}_{\pm 0.2}$ & $35.6_{\pm 1.8}$ & $35.0_{\pm 2.0}$ & $25.0_{\pm 2.3}$ & $25.0_{\pm 2.3}$ & $65.8_{\pm 0.2}$ & $26.1_{\pm 0.5}$ & $67.9_{\pm 0.3}$ & $66.3_{\pm 0.3}$ \\
~ & ISD & $81.7_{\pm 0.9}$ & $80.8_{\pm 0.9}$ & $44.2_{\pm 0.4}$ & $43.2_{\pm 0.5}$ & $29.5_{\pm 2.3}$ & $29.5_{\pm 2.3}$ & ${67.1}_{\pm 1.2}$ & $26.1_{\pm 0.1}$ & $68.0_{\pm 1.3}$ & $66.5_{\pm 1.2}$ \\
\cline{2-12}
~ & GatorTron & $\underline{89.4}_{\pm 0.5}$ & $\underline{88.4}_{\pm 0.5}$ & $66.0_{\pm 0.4}$ & $64.7_{\pm 0.7}$ & $\underline{57.1}_{\pm 2.2}$ & $\underline{57.1}_{\pm 2.2}$ & $75.5_{\pm 0.6}$ & $35.4_{\pm 1.9}$ & $77.3_{\pm 0.6}$ & $75.4_{\pm 0.6}$\\
~ & BioMedLM & $89.3_{\pm 0.3}$ & $88.2_{\pm 0.3}$ & $\underline{71.3}_{\pm 0.7}$ & $\underline{70.1}_{\pm 0.6}$ & $47.6_{\pm 4.3}$ & $47.6_{\pm 4.3}$ & $\underline{76.9}_{\pm 0.7}$ & $\underline{43.1}_{\pm 1.7}$ &	$\underline{78.4}_{\pm 0.6}$ & $\underline{76.6}_{\pm 0.6}$\\
\cline{2-12}
~ & DKEC-M-CNN & $85.2_{\pm 0.7}$ & $83.0_{\pm 0.7}$ & ${53.2}_{\pm 1.3}$ & ${52.7}_{\pm 1.1}$ & ${45.1}_{\pm 2.1}$ & ${45.1}_{\pm 2.1}$ & ${68.6}_{\pm 0.4}$ & ${32.4}_{\pm 0.6}$ & ${72.4}_{\pm 0.4}$ & ${71.7}_{\pm 0.6}$ \\
~ & DKEC-GatorTron & $\mathbf{91.8}_{\pm 0.1}$ & $\mathbf{90.7}_{\pm 0.1}$ & $\mathbf{72.4}_{\pm 0.4}$ & $\mathbf{71.3}_{\pm 0.4}$ & $\mathbf{67.6}_{\pm 2.3}$ & $\mathbf{67.6}_{\pm 2.3}$ & $\mathbf{79.5}_{\pm 0.5}$ & $\mathbf{51.1}_{\pm 1.5}$ & $\mathbf{82.2}_{\pm 0.5}$ & $\mathbf{80.3}_{\pm 0.6}$\\

\hline \hline

\multicolumn{2}{|c|}{} & P@8 & R@8 & P@8 & R@8 & P@8 & R@8 & miF & maF & P@8 & R@8\\
\cline{1-12}
\multirow{8}{*}{\rotatebox{90}{MIMIC-III}} & CAML & $54.8_{\pm 0.5}$ & $57.5_{\pm 0.6}$ & $5.5_{\pm 0.4}$ & $28.4_{\pm 2.3}$ & $0.7_{\pm 0.1}$ & $4.8_{\pm 0.5}$ & $51.5_{\pm 0.7}$ & $4.3_{\pm 0.5}$ & $54.4_{\pm 0.5}$ & $50.3_{\pm 0.5}$ \\


~ & ZAGCNN & $55.3_{\pm 0.2}$ & $58.0_{\pm 0.2}$ & $6.6_{\pm 0.1}$ & $34.4_{\pm 0.7}$ & $1.8_{\pm 0.1}$ & $11.7_{\pm 0.8}$ & $52.1_{\pm 0.4}$ & $4.0_{\pm 0.3}$ & $55.2_{\pm 0.2}$ & $51.2_{\pm 0.3}$ \\

~ & MultiResCNN & $\underline{56.5}_{\pm 0.3}$ & $\underline{59.4}_{\pm 0.2}$ & $\underline{8.2}_{\pm 0.5}$ & $\underline{42.3}_{\pm 2.8}$ & $1.2_{\pm 0.1}$ & $7.5_{\pm 0.9}$ & $\mathbf{55.6}_{\pm 0.3}$ & $\textbf{6.0}_{\pm 0.6}$ & $\underline{56.6}_{\pm 0.2}$ & $\underline{52.7}_{\pm 0.2}$ \\

~ & ISD & $51.8_{\pm 0.5}$ & $53.8_{\pm 0.5}$ & $6.1_{\pm 0.2}$ & $31.7_{\pm 1.2}$ & $1.9_{\pm 0.2}$ & $12.6_{\pm 0.9}$ & $46.8_{\pm 1.3}$ & $2.8_{\pm 0.2}$ & $51.6_{\pm 0.5}$ & $47.5_{\pm 0.5}$ \\

\cline{2-12}
~ & GatorTron & $50.4_{\pm 0.2}$ & $53.4_{\pm 0.2}$ & $6.5_{\pm 0.2}$ & $33.8_{\pm 1.1}$ & $2.0_{\pm 0.3}$ & $12.7_{\pm 1.4}$ & $45.4_{\pm 0.4}$ & $2.7_{\pm 0.3}$ & $50.3_{\pm 0.2}$	& $47.1_{\pm 0.2}$\\
~ & BioMedLM & $50.5_{\pm 0.1}$ & $53.4_{\pm 0.1}$ & $6.1_{\pm 0.1}$ & $31.3_{\pm 1.2}$ & $\underline{2.0}_{\pm 0.1}$ & $\underline{13.2}_{\pm 1.1}$ & $46.6_{\pm 0.3}$ & $3.7_{\pm 0.5}$ & $50.2_{\pm 0.1}$& $47.2_{\pm 0.2}$\\
\cline{2-12}

~ & DKEC-M-CNN & $\mathbf{58.6}_{\pm 0.2}$ & $\mathbf{61.5}_{\pm 0.2}$ & $\mathbf{9.6}_{\pm 0.1}$ & $\mathbf{49.2}_{\pm 0.8}$ & $2.9_{\pm 0.1}$ & $\mathbf{19.2}_{\pm 0.9}$ & $\underline{55.0}_{\pm 0.3}$ & $4.9_{\pm 0.2}$ & $\mathbf{58.9}_{\pm 0.2}$ & $\mathbf{54.8}_{\pm 0.2}$ \\
~ & DKEC-GatorTron & $56.8_{\pm 0.4}$ & $59.8_{\pm 0.2}$ & $8.5_{\pm 0.1}$ & $44.7_{\pm 0.7}$ & $\mathbf{3.1}_{\pm 0.2}$ & $19.1_{\pm 1.1}$ & $53.0_{\pm 0.4}$ &	$\underline{5.7}_{\pm 0.3}$ & $56.9_{\pm 0.4}$ & $53.2_{\pm 0.3}$\\

\hline
\end{tabular}
\vspace{-0.1in}
\caption{Comparison with SOTA on EMS and MIMIC-III. The best and runner-up results are in \textbf{bold} and \underline{underlined}.}
\label{sota comparison}
\vspace{-1em}
\end{table*}

\subsection{Datasets}
We used two datasets: a real-world EMS dataset, which is a collection of 4,417 pre-hospital electronic Patient Care Reports (ePCR) annotated with EMS protocol labels, and the benchmark EHR dataset, MIMIC-III~\cite{johnson2016mimic}, which is annotated with ICD-9 diagnosis codes. Both datasets contain 
textual descriptions of diagnoses, treatment protocols, interventions, and patient's medical history. Following the pre-processing steps in ~\cite{kim2021information}, we extract the relevant information from 4,417 ePCRs in EMS dataset. We use scikit-multilearn~\cite{szymanski2017scikit} to create 70:30 train/test splits for the EMS dataset and use 10\% of the train set for validation. 
Following the method in~\cite{mullenbach-etal-2018-explainable}, we split the train, validation, and test sets from the MIMIC-III but only consider a subset of 3,737 (out of 6,668) ICD-9 diagnosis codes as labels, since only for 3,737 of the codes we found knowledge available in the Wikipedia and Mayo Clinic websites. 
We separate the labels into three categories based on their frequencies in the dataset: \textit{head labels ({H})} with more than 1,000 samples, \textit{middle labels ({M})} with 10 to 100 samples, and \textit{tail labels ({T})} with less than 10 samples (few-shot cases). Table~\ref{tab:data statistics} shows statistics of the datasets.

\subsection{Knowledge Bases}
\label{kb_description}
We constructed two separate heterogeneous graphs for capturing the domain knowledge for ICD-9 diagnosis codes in MIMIC III and protocols in the EMS dataset. For ICD-9 diagnosis codes, \textbf{Wikipedia} and \textbf{Mayo Clinic} web contents are scraped. For EMS protocols, we use symptom and procedure sections from official EMS guidelines, which are available on the Old Dominion EMS Alliance (\textbf{ODEMSA}) website~\footnote{https://odemsa.net/}. Statistics of two knowledge graphs are in Appendix~\ref{KG_statistics}.
To evaluate the accuracy of different methods for constructing knowledge graphs, we evenly sampled 50 codes from head, middle, and tail classes and manually annotated symptoms and treatments from Wikipedia and Mayo Clinic website contents for ICD-9 diagnosis codes. For EMS protocols, we manually annotated all 43 protocols in ODEMSA documents. The extracted web contents, ground truth annotations, and knowledge graphs are here~\footnote{https://github.com/UVA-DSA/DKEC}.

Both rule-based and ML-based methods are used as the medical entity extraction baselines, including MetaMap~\cite{aronson2001effective}, cTAKES~\cite{savova2010mayo}, ScispaCy~\cite{neumann-etal-2019-scispacy}. The prompt templates for zero-shot and one-shot CoT and baseline model configurations are shown in Appendix~\ref{sec:NER_model_specification}. Micro f1-score for entity extraction is reported. We count an extracted medical entity as correct if it exactly matches the ground truth.

\subsection{Metrics and Parameter Settings}
We report the micro F1 ($miF$) and macro F1 ($maF$) scores with a fixed threshold of 0.5. $miF$ is heavily influenced by frequent diagnosis codes and thus can be used to evaluate the performance of the head/middle classes. $maF$ weighs the F1 achieved on each label equally and is used to evaluate the performance for the tail classes. Ranking-based metrics~\cite{chalkidis-etal-2019-large, chalkidis2020empirical} recall at $k$ ($R@K$) and precision at $k$ ($P@K$), which do not require a specific threshold, are also reported. $P@K$ is important as it measures the proportion of relevant diagnosis codes suggested in top-k recommendations by the model. $R@K$ is important for medical professionals when they consider the most probable diagnoses for treatments. As the average number of labels per instance in MIMIC-III is 8.0 and EMS is 1.2, we set $K$ as 8 and 1, respectively. 

To reduce noise, we did a pre-processing step to remove punctuations and stopwords. We trained each model 5 times, each time with a different random initialization seed. We report the mean $\pm$ standard deviation of results 
with the best parameters. The hidden state size and number of attention heads in graph models are set as 256 and 8, respectively. 
We use Adam optimizer and regularization with a weight decay of 1e-5 and a dropout rate of 0.2. For training the baselines, we use their best parameter settings. We developed all models by PyTorch~\cite{paszke2019pytorch} and Huggingface~\cite{wolf2019huggingface}. All experiments were run on NVIDIA GPU A100 (more details are in Appendix~\ref{sec:train_details}). 

\subsection{Baselines}
We evaluate DKEC in comparison to the following SOTA networks for diagnosis prediction: \\
\textbf{Pre-trained Transformers}: 
BERT models including \textbf{TinyClinicalBERT}(15M), \textbf{DistilBioBERT}(66M)~\cite{rohanian2023lightweight}, \textbf{COReBERT}(110M) and LLMs like \textbf{GatorTron}(325M) and \textbf{BioMedLM}(2.7B) are pre-trained on external biomedical knowledge for clinical NLP tasks.\\
\textbf{Label-wise Attention Networks}: The following baselines were selected due to their superior performance and code availability~\cite{ji2022unified}:\\
\textbf{CAML}: The convolutional attention network for multi-label classification~\cite{mullenbach-etal-2018-explainable} learns attention distribution for each label. \\
\textbf{ZAGCNN}: Zeroshot attentive GCNN ~\cite{rios2018few} integrates hierarchical structure of ICD codes by graph CNNs to select label-relevant features for ICD classification. \\
\textbf{MultiResCNN}: Multi-Filter Residual CNN~\cite{li2020icd} utilizes a multi-filter convolutional layer to capture n-gram patterns and a residual mechanism to enlarge the receptive field.\\ 
\textbf{ISD}: Interactive shared representation network with self-distillation~\cite{zhou2021automatic} models connections among labels and their co-occurrence.

\section{Experimental Results}
\subsection{Knowledge Graph Quality Evaluation}
\label{KG_quality_eval}
As shown in Table~\ref{KG evaluation}, \textit{\textbf{one-shot CoT GPT-4 outperforms other baselines in medical entity extraction}} consistently by a considerable margin. Zero-shot GPT-4 has better performance in extracting symptoms than treatments. In our detailed manual evaluations, we observed that zero-shot GPT-4 usually outputs the whole sentence containing a medical entity or rephrases the medical entities during extraction (see the example in Appendix~\ref{zero-shot}). With token classification and span detection in one-shot CoT GPT-4 we avoid this problem. 

\subsection{Class Imbalance Analysis}
Table~\ref{sota comparison} shows the performance of DKEC when applied to Multi-filter CNN (DKEC-M-CNN) and GatorTron (DKEC-GatorTron) vs. SOTA on EMS and MIMIC-III datasets. For all the head/middle/tail classes, DKEC outperforms all the baselines. 
Several observations are highlighted:


\textit{\textbf{DKEC alleviates the class imbalance problem}}. As shown in Table~\ref{sota comparison}, improvement is most evident on the tail labels. 
DKEC achieves 10.5\% and 6\% increase in top-k recall on EMS and MIMIC-III datasets, respectively, compared with runner-up SOTA. On the middle labels
, the improvement is still considerable. Compared with runner-up SOTA, DKEC achieves a 6.9\% improvement in top-k recall on MIMIC-III. In the head labels where there are sufficient samples, the improvement is relatively small ($\sim$2\%). 
Overall, DKEC maintains a comparable performance to baselines for the head labels while achieving better performance for middle and tail labels, which narrows down the performance gap regardless of data distribution (\textbf{RQ1}). We also do an error analysis to understand where and why DKEC underperforms in Appendix~\ref{error_analysis}.

We also observe that transformer models in general achieve a lower performance compared to CNN models on the MIMIC-III dataset, but outperform them on the EMS dataset. This may be due to the different characteristics of the datasets. The EMS dataset contains fewer training samples and labels per sample, while the MIMIC-III notes are longer and each contain a larger number of labels. One hypothesis is that pre-trained transformers perform better on shorter notes and with fewer training samples, which can be further studied in future work.

\subsection{Model Size vs. Performance}
LLMs have great few-shot abilities but they are costly to train and deploy on resource-constrained devices~\cite{jin2023emsassist, weerasinghe2024real}. So, we apply DKEC to transformers of varying sizes, including LLMs, to answer \textbf{RQ2}.

\begin{figure}[t!]
	\centering
	\begin{minipage}{1\linewidth}	
		\subfigure[EMS]{
			\label{fig:a}
			\includegraphics[width=0.49\linewidth,height=1in]{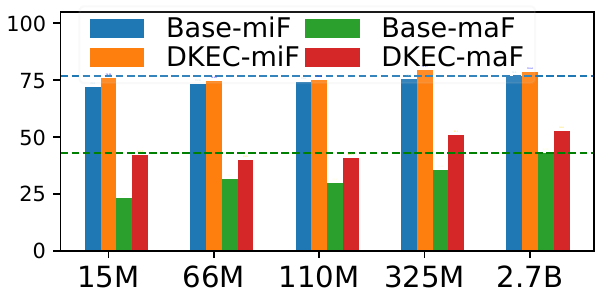}
		}\noindent
		\subfigure[MIMIC-III]{
			\label{fig:b}
			\includegraphics[width=0.49\linewidth,height=1in]{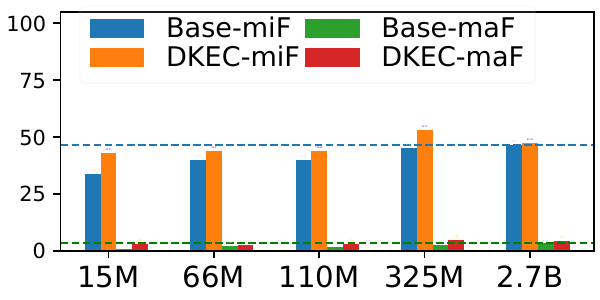}
		}
	\end{minipage}
    \vspace{-1.3em}
	\caption{DKEC with different pre-trained transformers.
    }
	\label{comparison over models}
    \vspace{-1.4em}
\end{figure}

\textit{\textbf{Performance of DKEC-based models increase less as model size grows}}. Our results show that DKEC is \textit{model-agnostic} and can be applied to different model architectures and sizes from 15M to 2.7B. However, as shown in Figure~\ref{fig:a} and~\ref{fig:b}, DKEC has more improvement on small language models than LLMs. For example, when applying DEKC, there is a 18.8\% improvement in $maF$ over TinyClinicalBERT (15M), while there is only a 9.5\% improvement over BioMedLM (2.7B) on the EMS dataset. This might be because LLMs are pre-trained on extensive medical corpora and can handle longer texts, thus show better few-shot abilities. \textit{\textbf{DKEC enables smaller language models to achieve comparable performance to LLMs.}}  As shown in Figures~\ref{fig:a} and~\ref{fig:b}, in both datasets, GatorTron (325M) with DKEC outperforms baseline BioMedLM (2.7B) in both $miF$ and $maF$. This indicates the benefit of DKEC in enabling less costly deployment of small language models in real-world applications.

\subsection{Label Size vs. Performance}
To understand the effect of label size on the performance of DKEC (\textbf{RQ3}), we conduct experiments on subsets of MIMIC-III dataset with varying label sizes, including 1.0k, 3.7k, and 6.7k labels. 
The knowledge from online sources are fully available for subsets with 1.0k and 3.7k labels, while partially available for the full dataset with 6.7k labels. 

\begin{figure}[t!]
    \centering
    \includegraphics[width=0.9\linewidth,height=2.7cm]{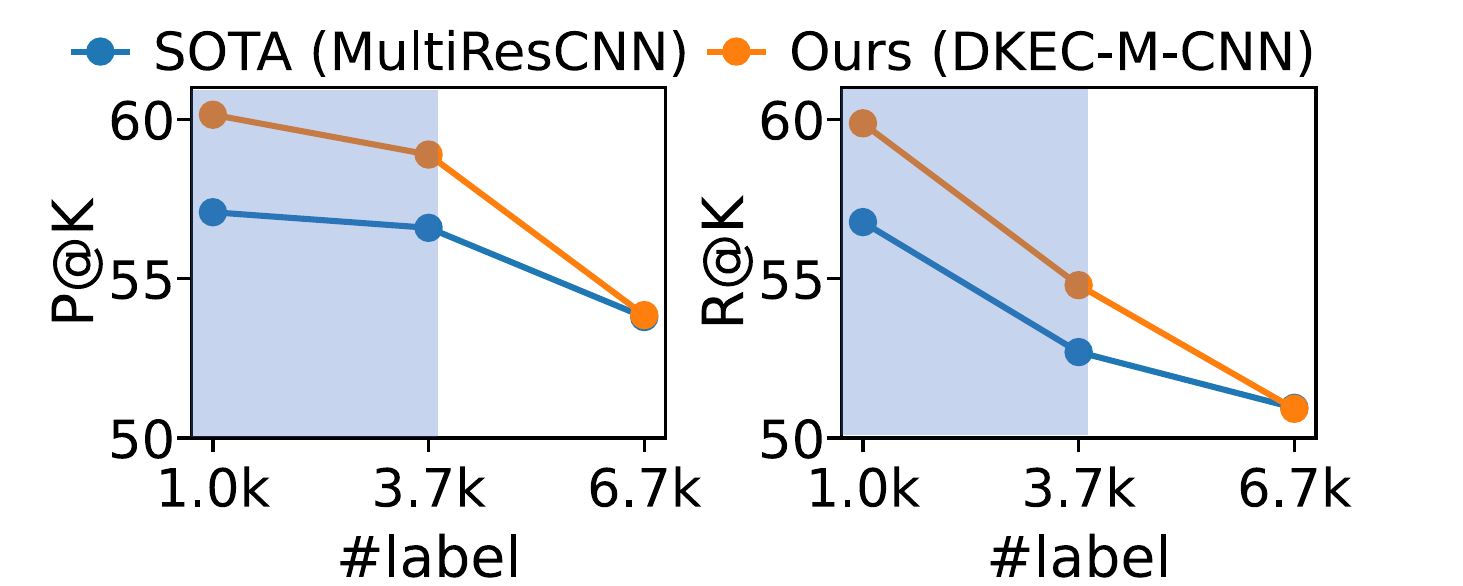}
    \vspace{-0.1in}
    \caption{Performance on subsets of MIMIC-III dataset with varying label sizes. Subsets with 1.0k and 3.7k labels have full knowledge. 6.7k has partial knowledge.}
    \label{label comparison}
 \vspace{-1.4em}
\end{figure}

As shown in Figure~\ref{label comparison}, with 6.7k labels, DKEC-M-CNN has similar performance to the best SOTA only with partial knowledge. However, on datasets with 3.7k and 1.0k labels (with full knowledge), DKEC-M-CNN outperforms the best SOTA MultiResCNN by 2$\sim$4\% with some memory cost during inference ($\sim$200MB). 
\textit{\textbf{With the increase in the number of labels, the MLTC performance generally drops, but DKEC helps maintain performance, particularly when external knowledge is available for all the labels.}} 
More results on comparison of DKEC with SOTA with 6.7k and 1.0k labels are available in Appendix~\ref{6668-1000}.

\subsection{Ablation Study}
To investigate the effectiveness of DKEC, we conduct an ablation study on the effect of different encoders and label-wise attention mechanisms using MIMIC-III dataset, as shown in Table~\ref{ablation study}. 
\\
\textbf{Effectiveness of DKEC:} For both encoders, incorporating DKEC leads to better performance compared to only using label hierarchy which is the SOTA label-wise attention mechanism.\\
\textbf{Effectiveness of External Knowledge:} When applying DKEC to M-CNN without hierarchical label relations, performance shows minimal fluctuations. This suggests that incorporating label-specific semantic relations from external knowledge sources is the main driver of performance improvements.

\begin{table}[h!]
\vspace{-0.4em}
\small
\centering
\setlength{\tabcolsep}{3 pt} 
\begin{tabular}{|c|c|cccc|} 
\hline
Encoder & Label-wise Attention  & miF & maF & P@8 & R@8\\
\cline{1-6}
1-CNN & Label hierarchy\textsuperscript{*} & $52.1$ & $4.0$ & $55.2$ & $51.2$\\
1-CNN & DKEC & $54.8$ & $4.2$ & $57.5$ & $53.3$\\
GatorTron & Label hierarchy & $46.6$ & $3.2$ & $50.7$ & $47.5$\\
GatorTron & DKEC & $53.0$ & \textbf{5.7} & 56.9 & 53.2  \\
M-CNN & DKEC & 55.0 & 4.9 & \textbf{58.9} & \textbf{54.8}\\
M-CNN & DKEC w/o hierarhcy  & \textbf{55.2} & 4.9 & 58.6 & 54.5\\
\hline
\end{tabular}
\vspace{-0.05in}
\caption{Ablation study using MIMIC-III dataset. ``1/M-CNN'' are the single/multi-filter CNN. 1-CNN with Label hierarchy$^{\star}$ represents the SOTA ZAGCNN.} 
\label{ablation study}
\vspace{-0.5em}
\end{table}

\section{Conclusion}
This paper proposes a domain knowledge-enhanced multi-label text classification method for diagnosis prediction. We present an approach for automatic knowledge graph construction from online sources based on chain-of-thought prompting with GPT-4. We also introduce a heterogeneous label-wise attention that incorporates relations among diverse medical entities in the knowledge graph to capture label-related text features for classification. We evaluated our methods on two real-world datasets. Experiments show the accuracy of knowledge graph construction based on three knowledge bases and improved performance over SOTA methods. We also demonstrated the applicability of our approach to different language models sizes and its scalability to large number of labels.

\section*{Limitations}
Firstly, we only construct the heterogeneous graphs using two KBs, Wikipedia and Mayo Clinic.  This leads to extracting relevant domain knowledge for only a subset of 3,737 diagnosis codes in the  MIMIC-III dataset. Larger KBs might be needed to build more complete knowledge graphs. We do not use UMLS as a knowledge base because UMLS interface does not provide direct heterogeneous relations between “disease” and “sign or symptoms” (subset of Finding) and “disease” and “treatments”, which are required for constructing the heterogeneous graphs in this work. Besides, constructing knowledge graphs by UMLS might require extensive manual effort. 

Secondly, we manually annotated 50 ICD9 diagnosis codes to illustrate the accuracy of different methods for medical entity extraction. The accuracy of the full knowledge graph would require a considerable amount of human effort.

Lastly, although DKEC performs better than SOTA, more efforts are needed to improve the prediction accuracy of rare diagnosis codes. Given the current SOTA prediction accuracy, particularly for ICD-9 codes in the MIMIC-III dataset, the predictions by DKEC and other baselines studied in this paper should be only used as a reference by healthcare workers and not as a final decision for treatment of the patients in the real world.  Further human evaluation of diagnosis prediction models and feedback from medical experts with extensive knowledge of diseases and patient conditions are needed. Large-scale human evaluation and understanding of how the top-k recall and precision results translate to human trust in the system requires more research efforts, which are beyond the scope of this paper.



\bibliographystyle{acl_natbib}
\bibliography{custom}

\appendix
\section{Appendix}
\label{sec:appendix}

\subsection{UMLS Normalization}
\label{sec:umls_norm}
We call UMLS API to get top10 ranked medical concepts for a medical entity, and return the first result with a relevant semantic type from our pre-defined set of semantic types for symptoms and treatments. The algorithm is as follows,
\begin{algorithm}
\caption{Medical Entity Normalization}\label{alg:med_norm}
\renewcommand{\algorithmicrequire}{\textbf{Input:}}
\renewcommand{\algorithmicensure}{\textbf{Output:}}
\begin{algorithmic}[1]
    \Require Pre-defined Semantic Set $S$, $k$th medical entity $e_{k}$, UMLS API
    \Ensure Normalized Concept $e^{norm}_{k}$
    \State $r_{k=1}^{10}=$ UMLS($e_{k}$) \Comment{$r_k$ includes normalized concept, CUI, semantic types and etc.}

    \For{$k \gets 1$ to $10$}
        \If{Semantic type of $r_k$ in $S$}
            \State $e^{norm}_{k} =$ normalized concept of $r_k$
            \State \Return $e^{norm}_{k}$
        \EndIf
    \EndFor
    \State \Return \textbf{null}
\end{algorithmic}
\end{algorithm}

We provide the relevant semantic type set for medical concept normalization.\\
For \textit{Signs and Symptoms}, the relevant semantic types are "Sign or Symptom (sosy)", "Disease or Syndrome (dsyn)", "Mental or Behavioral Dysfunction (mobd)", "Neoplastic Process (neop)", "Anatomical Abnormality (anab)", "Finding (fndg)", "Pathologic Function (patf)", "Congenital Abnormality (cgab)", and "Injury or Poisoning (inpo)."

For \textit{Treatment}, they are "Therapeutic or Preventive Procedure (topp)", "Antibiotic (antb)", "Clinical Drug (clnd)", "Vitamin (vita)", "Organic Chemical (orch)", and "Amino Acid, Peptide, or Protein (aapp)", "Pharmacologic Substance (phsu)", "Laboratory Procedure (lbpr)", and "Diagnostic Procedure (diap)."

\subsection{Knowledge Graph Statistics}
\label{KG_statistics}
The detailed statistics of knowledge graph built for MIMIC-III (3737 labels) and EMS (43 labels) datasets are shown in Table~\ref{tab:Graph_Metrics}. Wikipedia contributes most of unique nodes and edges to the union graph (WoU), and Mayo Clinic website complements some other unique nodes (MoU).

\begin{table*}[htbp]
\centering
\small
    \begin{tabular}{|c|c|c|c|c|c|}
    \hline
    Dataset & \multicolumn{4}{|c|}{MIMIC-III} & EMS\\
    \hline
    \multirow{2}{*}{Knowledge Graph} & \multirow{1}{*}{Wikipedia $W$} & \multirow{1}{*}{Mayo Clinic $M$ } & \multicolumn{2}{c|}{Combined} & \multirow{2}{*}{ODEMSA}\\
         & (WoU) & (MoU) &$W \cup M$ & $W \cap M$ (IoU)  & \\
    \hline
    \# Total Nodes & 19275 (0.93) & 3672 (0.18) & 20835  &2112 (0.10) & 497\\
    \# Disease Nodes & 3666 (0.98) & 320 (0.09)& 3737 & 249 (0.07)& 43 \\
    \# Sign and Symptom Nodes & 7900 (0.90)& 1800 (0.20) & 8805 & 895 (0.10)  & 310\\
    \# Treatment Nodes & 7709 (0.93)& 1552 (0.19)& 8293 & 968 (0.12) & 144 \\
    \# Sign and Symptom Edges & 37273 (0.92) & 3738 (0.09) & 40566 & 445 (0.01) & 610\\
    \# Treatment Edges & 42267 (0.94) & 3068 (0.07) & 44785 & 550 (0.01) & 328 \\
    \hline
    \end{tabular}
    \vspace{-0.1in}
    \caption{Knowledge Graph Statistics.
    WoU = \(\frac{|W|}{|W \cup M|}\);
    MoU = \(\frac{|M|}{|W \cup M|}\);
    IoU = \( \frac{|W \cap M|}{|W \cup M|}\).
    }
    \vspace{-0.1in}
    \label{tab:Graph_Metrics}
\end{table*}

\subsection{Medical Entity Extraction Model Configurations}
\label{sec:NER_model_specification}
We used the following configurations for MetaMap, cTAKES, ScispaCy and GPT-4.


For MetaMap, we use the Python wrapper pymetamap\footnote{https://github.com/AnthonyMRios/pymetamap} to do the named entity recognition, the hyper-parameters in pymetamap are the semantic types defined in Appendix~\ref{sec:umls_norm}. For cTAKES, we used the DefaultFastPipeline.piper configuration and then filtered for "textsem: SignSymptomMention". For ScispaCy, we used the python interface of the model \verb|en_core_sci_scibert|\footnote{https://allenai.github.io/scispacy/} to extract entities and filter by the semantic types defined in Appendix~\ref{sec:umls_norm}.

For zero-shot GPT-4 and one-shot CoT GPT-4, we provide one example of prompting and its response.\\
\textbf{Zero-shot GPT-4}: The zero-shot GPT-4 sometimes outputs the whole sentence containing the symptoms or treatments or rephrases them, as shown in the example below.
\begin{tcolorbox}[breakable]
\label{zero-shot}
\textbf{PROMPT: }Given a biomedical text, extract the symptoms and treatments for the disease. The symptom includes signs and symptoms of the disease. The treatment includes medications and procedures used to cure the disease. Remove the dosage of medications.\par
Return your result in latex format as follows,\par
\begin{verbatim}
```
\begin{SYMPTOM}
\item{Symptom_1}
...
\item{Symptom_n}
\end{SYMPTOM}
\begin{TREATMENT}
\item{Treatment_1}
...
\item{Treatment_n}
\end{TREATMENT}
```
\end{verbatim}
[...]\par
\textbf{RESPONSE: }\par
\begin{verbatim}
```
\begin{SYMPTOM}
\item{Watery diarrhea three or more 
times a day for more than one day}
\item{Mild belly cramping and 
tenderness}
[...]
\item{Toxic megacolon}
\item{Sepsis}
\end{SYMPTOM}
\begin{TREATMENT}
\item{Antibiotics}
\item Vancomycin (Firvanq Kit)
\item Fidaxomicin (Dificid)
\item Metronidazole (Flagyl), 
used in combination with vancomycin 
for severe infection
[...]
\item Probiotics - being researched 
for treatment or prevention, not 
confirmed for effectiveness outside 
of studies
\end{TREATMENT}
```
\end{verbatim}

\end{tcolorbox}

\textbf{One-shot CoT GPT-4}
\begin{tcolorbox}[breakable]
\textbf{PROMPT: }Given a biomedical text, extract the symptoms and treatments for the disease. Follow the four steps in the following example to generate the results. Pay attention to the requirements in every steps. Return your result in latex format.\par
\begin{verbatim}
```
\begin{SYMPTOM}
\item{Symptom_1}
...
\item{Symptom_n}
\end{SYMPTOM}
\begin{TREATMENT}
\item{Treatment_1}
...
\item{Treatment_n}
\end{TREATMENT}
```
\end{verbatim}

Here is one example: the disease is Emphysema\par
TEXT: Early symptoms include cough, wheezing, shortness of breath. Other symptoms may include anxiety, depression. Give antibotics if having bacterial infection. Lung transplantation is an option if having severe lung damage. Do not use antibotic medications like Metronidazole (Flagyl)\newline

Response: Let's think step by step,\par
Step1: label the tokens one by one "symptom", "treatment" or "none", The “symptom” event type includes signs and symptoms. The “treatment” event type includes medicine and clinical procedures. The "none" event type indicates the token does not belong to "symptom" and "treatment".\par
\begin{verbatim}
-Early: none
[...]
-Flagyl: treatment
\end{verbatim}\newline
Step2: Refine symptoms and treatments from Step 1 by following criteria,\par
1.concatenate symptom/treatments spans\par
2.remove extra irrelevant words in symptoms and treatments\par
\begin{verbatim}
-cough: symptom
-wheezing: symptom
-shortness of breath: symptom
-anxiety: symptom
-depression: symptom
-antibotics: treatment
-Lung transplantation: treatment
-Metronidazole: treatment
-Flagyl: treatment
\end{verbatim}

Step3: identify the relation [positive, negative] between extracted symptoms/treatments from Step 2 with the disease according to the provided content. Positive means extracted phrase is the symptom of the disease/treatment improve the disease, negative means extracted phrase is not a symptom of the disease/treatment worsen the disease.
\begin{verbatim}
-cough: positive
-wheezing: positive
-shortness of breath: positive
-anxiety: positive
-depression: positive
-antibotics: positive
-Lung transplantation: positive
-Metronidazole: negative
-Flagyl: negative
\end{verbatim}

Step 4: take symptoms/treatments with positive relations with disease from Step 3 and reorganize the result \
as latex format. Do not include explanation, conditions from Step 3 in your results.
\begin{verbatim}
```
\begin{SYMPTOM}
\item{cough}
\item{wheezing}
\item{shortness of breath}
\item{anxiety}
\item{depression}
\end{SYMPTOM}
\begin{TREATMENT}
\item{antibotics}
\item{Lung transplantation}
\end{TREATMENT}
```
\end{verbatim}

Now is the real task: the disease is \{title\}\par
TEXT: [...]\par
Response: Let's think step by step\newline

\textbf{RESPONSE:}
\end{tcolorbox}

\begin{table*}[htbp]
\small
\centering
\setlength{\tabcolsep}{4 pt} 
\begin{tabular}{|c|c|cc|cc|cc|cccc|} 
\hline
\multicolumn{2}{|c|}{} & \multicolumn{2}{c|}{Head Labels}& \multicolumn{2}{c|}{Middle Labels} & \multicolumn{2}{c|}{Tail Labels} & \multicolumn{4}{c|}{Overall}\\

\multicolumn{2}{|c|}{} & P@6 & R@6 & P@6 & R@6 & P@6 & R@6 & miF & maF & P@6 & R@6\\
\cline{1-12}

\cline{3-12}
\multirow{8}{*}{\rotatebox{90}{MIMIC-III-1000}} & CAML & $56.1$ & $58.2$ & $6.8$ & $35.8$ & ${2.9}$ & ${17.7}$ & $56.0$ & $8.3$ & $55.8$ & $55.1$\\
~&ZAGCNN &$56.4$ & $58.9$ & $7.4$ & $38.8$ & $2.1$ & $12.8$ & $55.9$ & $9.0$ & $56.2$ & $55.7$\\
~&MultiResCNN & $\underline{57.2}$ & $\underline{59.9}$ & $\underline{9.1}$ & $\underline{47.6}$ & $2.6$ & $15.7$ & $\underline{59.4}$ & $\underline{11.5}$ & $\underline{57.1}$ & $\underline{56.8}$\\
~&ISD & $54.5$ & $56.2$ & $6.9$ & $36.2$ & $2.0$ & $11.8$ & $54.6$ & $6.9$ & $54.2$ & $53.2$ \\

\cline{2-12} 
~ & GatorTron & $52.6$ & $54.9$ & $8.4$ & $43.8$ & ${3.8}$ & ${22.6}$ & $50.9$ & $6.1$ & $52.4$ & $51.9$\\
~ & BioMedLM & $54.3$ & $57.1$ & $8.9$ & $46.7$ & $\underline{5.4}$ & $\underline{32.4}$ & $53.4$ & $7.9$ &	$54.3$	& $54.3$\\
\cline{2-12}

~&DKEC-M-CNN & $\mathbf{60.2}$ & $\mathbf{62.9}$ & ${11.4}$ & ${59.2}$ & ${4.6}$ & ${27.5}$ & $\mathbf{61.5}$ & ${12.2}$ & $\mathbf{60.2}$ & $\mathbf{59.9}$\\
~&DKEC-GatorTron & $57.6$ & $60.5$ & $\mathbf{11.5}$ & $\mathbf{60.1}$ & $\mathbf{6.5}$ & $\mathbf{39.2}$ & $57.7$ & $\mathbf{13.9}$ & $57.7$ & $57.8$\\

\hline
\hline
\multicolumn{2}{|c|}{} & P@12 & R@12 & P@12 & R@12 & P@12 & R@12 & miF & maF & P@12 & R@12\\
\hline
\multirow{8}{*}{\rotatebox{90}{MIMIC-III-6668}} & CAML & $51.3$ & $55.9$ & $4.7$ & $28.1$ & $0.5$ & $4.0$ & $46.4$ & $3.6$ & $51.2$ & $48.2$\\
~&ZAGCNN & $51.3$ & $55.9$ & $6.1$ & $36.0$ & $1.4$ & $12.1$ & $47.6$ & $3.9$ & $51.5$ & $48.8$\\
~&MultiResCNN & $\underline{53.2}$ & $\underline{58.1}$ & $\mathbf{7.6}$ & $\mathbf{44.9}$ & $1.1$ & $9.3$ & $\mathbf{51.4}$ & $\mathbf{6.2}$ & $\underline{53.8}$ & $\mathbf{51.0}$\\
~&ISD & $46.3$ & $50.2$ & $5.2$ & $31.0$ & $\underline{1.6}$ & $\underline{13.9}$ & $39.0$ & $2.5$ & $46.0$ & $43.5$\\
\cline{2-12}
~&GatorTron & $43.4$ & $47.9$ & $5.1$ & $30.8$ & $1.2$ & $9.3$ & $37.1$ & $1.8$ & $43.6$ & $41.8$\\

~ & BioMedLM & $44.3$ & $48.9$ & $4.9$ & $29.1$ & $1.2$	& $9.7$	& $39.2$ & $2.4$ & $44.4$ & $42.7$\\

\cline{2-12}
~&DKEC-M-CNN &$\mathbf{53.5}$ & $\mathbf{58.3}$ & $\underline{6.9}$ & $\underline{41.2}$ & ${1.8}$ & $\mathbf{16.4}$ & $\underline{48.7}$ & $\underline{4.5}$ & $\mathbf{53.9}$ & $\underline{50.9}$\\
~&DKEC-GatorTron & $51.7$ & $56.9$ & $6.8$ & $41.0$ & $
\mathbf{1.9}$ & 	$15.3$ & $46.4$ & $4.3$ & $52.1$ & $49.8$\\

\cline{1-12}
\hline
\end{tabular}
\vspace{-0.1in}
\caption{Comparison with SOTA on MIMIC-III-1000 and MIMIC-III-6668 datasets. The best result is highlighted in \textbf{bold}, and the runner-up is \underline{underlined}.}
\label{mimic-6668-1000 comparison}
\vspace{-0.1in}
\end{table*}

\subsection{Training Implementation Details}
\label{sec:train_details}
For all models, we use the same early stopping criteria, namely, if the $R-Precision@K$ does not increase for more than 1e-3 for 10 times, the training is stopped. During the testing, we select the the epoch model that has the best performance. We set the maximum epoch as 200 for CNN-based models. For pre-trained transformers, we set the maximum as 30. This is because pre-trained transformers usually need less epochs to converge while CNN is trained from scratch.

In graph node initialization, we use BlueBERT~\cite{peng2019transfer} to generate the embedding for DKEC-M-CNN. For pre-trained transformers, we directly use themselves to generate the initial node embeddings. For EMS dataset, we set the HGT layer in graph model as 1 while in MIMIC3 dataset we set the HGT layer as 2. The five random seeds we used are 0, 1, 42, 1234, 3407.

For pre-trained transformers, max-pooling is used before the classification, but for DKEC-M-CNN models, sum-pooling is used instead. Experiments show sum-pooling works best for DKEC-M-CNN, and max-pooling works best for pre-trained transformers. One disadvantage of max-pooling is that it will consume more computation resources than sum-pooling, especially for datasets that have huge amounts of labels. However, there is no clear evidence on which pooling mechanism is optimal. One recommendation is to choose the pooling mechanism based on the data and the need for coding practice. 
For BioMedLM(2.7B), the FSDP and BFloat16 are applied to speed up training. The last token embedding is used as the document feature for classification.

\subsection{Performance on MIMIC-III with different label sizes}
\label{6668-1000}
As shown in Table~\ref{mimic-6668-1000 comparison}, we show the performance on MIMIC-III-6668 (all ICD-9 diagnosis codes), and MIMIC-III-1000 (sampled ICD-9 diagnosis codes). Note that For MIMIC-III-6668, only partial labels have domain knowledge while for MIMIC-III-1000, all labels have domain knowledge. And we run all models on the random seed of 3407. All the other settings are the same with as reported in implementation details~\ref{sec:train_details}.

On MIMIC-III-6668 where partial knowledge is available, DKEC-M-CNN has similar overall performance with the best SOTA, with tiny improvement on the tail labels and tiny decrease on the middle labels. On MIMIC-III-1000 where every label has external knowledge (partial knowledge), DKEC-M-CNN outperforms the best SOTA in overall performance and significantly improves the performance in the middle and tail labels.

\begin{table}[t!]
\small
\centering
\setlength{\tabcolsep}{3 pt} 
\begin{tabular}{|c|cc|cc|cc|} 
\hline
\multirow{2}{*}{Datasets} & \multicolumn{2}{|c|}{Head} & \multicolumn{2}{|c|}{Mid} & \multicolumn{2}{|c|}{Tail}\\
~ & FP & FN & FP & FN & FP & FN\\
\cline{1-7}
EMS & 199 & 128 & 107 & 124 & 4 & 18 \\
MIMIC-III & 4167 & 15110 & 131 & 2796 & 1 & 608\\
\hline

\hline
\end{tabular}
\vspace{-0.05in}
\caption{Numbers of FPs, FNs in head/middle/tail labels}
\label{error_analysis}
\vspace{-0.in}
\end{table}

\subsection{Error Analysis}
\label{error_analysis}
To investigate where DKEC underperforms, we do an error analysis by selecting the best model for each dataset (DKEC-M-CNN for MIMIC-III dataset, and DKEC-GatorTron for EMS dataset) and calculate the FP/FN on head/middle/tail labels. As shown in~\ref{error_analysis}, the number of FNs is much larger than FPs (FN > FP). This indicates that our model is very conservative in decision-making when there are fewer training samples. Further research is needed to alleviate the data imbalance problem.

By manually checking some examples (50 / 253) of model mis-predictions for the EMS dataset, we find two main reasons for the errors. First, as shown in Figure~\ref{eg:spurious_relation}, DKEC model is vulnerable to \textbf{spurious relations and have no causal reasoning ability} to differentiate the main sign/symptoms for a disease from the secondary ones. It shows the model’s inability to analyze the causation between signs/symptoms and diagnosis and just makes decision based on non-relevant words. Future work can focus on debiasing and causal reasoning for improving diagnosis prediction.\\
Secondly, DKEC is \textbf{not effective in distinguishing similar labels}, for example, As shown in Figure~\ref{eg:similar_labels}, in an EHR narrative “difficulty breathing” and “upper respiratory infection” symptoms are mentioned and the ground truth is “medical - respiratory distress/asthma/copd/croup/reactive airway” but the model mispredicted it as “airway - failed”. Both of the labels are respiratory system-related problems and confuse the model in decision-making. Similar observations can be found in the case of "childbirth" and "pre-term labor".

\begin{tcolorbox}[width=\linewidth,breakable]
\textbf{Example 1}: \par
\textbf{ePCR}: ...started to have \textcolor{red}{abdominal cramping} and \textcolor{blue}{vomiting}. The patient \textcolor{blue}{vomited twice}... \textcolor{red}{Abdomen}) Soft with \textcolor{red}{abdominal pain} to the lower quadrants. \textcolor{red}{10 out 10 cramping pain}...\par
\textbf{True Label}: “medical - abdominal pain” \par
\textbf{Prediction}: “medical - nausea/vomiting”, "medical - abdominal pain" \\

\textbf{Example 2}:\par
\textbf{ePCR}: ...The patient appeared to be working \textcolor{red}{hard to breath with fast rate}. C: \textcolor{red}{Difficulty Breathing}. H: The patient states that he was late for \textcolor{blue}{dialysis} yesterday and they did not take enough fluid off ...The patient denied any history of CHF, however did say he is \textcolor{blue}{diabetic}... \par
\textbf{True Label}: "airway - failed" \par
\textbf{Prediction}: "airway - failed", "medical - diabetic - hyperglycemia" 
\end{tcolorbox}
\noindent\begin{minipage}{\linewidth}
\vspace{-0.2in}
\captionof{figure}{Spurious relation examples. Label-related keywords are in \textcolor{red}{red}, and spurious words are in \textcolor{blue}{blue}.}\label{eg:spurious_relation}
\vspace{-0.3in}
\end{minipage}

\begin{tcolorbox}[width=\linewidth, breakable]
\textbf{Example 1}:\par
\textbf{ePCR}: ...The patient was in moderate distress with \textit{difficulty breathing}. The patient appeared to be anxious, and was \textit{breathing at fast rate}. C: \textit{shortness of breath}...\par
\textbf{True Label}: "medical - respiratory distress/asthma/copd/croup/reactive airway"\par
\textbf{Prediction}: "airway - failed", "medical - respiratory distress/asthma/copd/croup/reactive airway"\\

\textbf{Example 2}: \par
\textbf{ePCR}: ...Pt was told that she was having \textit{miscarriage}. At the time, pt was months \textit{pregnant}. Pt stated that tonight the \textit{bleeding got significantly worse} and that she has been passing \textit{large clots}...\par
\textbf{True Label}: "ob/gyn - childbirth/labor/delivery"\par
\textbf{Prediction}: "ob/gyn - pre-term labor"
\end{tcolorbox}
\noindent\begin{minipage}{\linewidth}
\vspace{-0.115in}
\captionof{figure}{Similar label examples}\label{eg:similar_labels}
\end{minipage}

\end{document}